\documentclass[letterpaper]{article} 
\usepackage[]{aaai25}  
\usepackage{times}  
\usepackage{helvet}  
\usepackage{courier}  
\usepackage[hyphens]{url}  
\usepackage{graphicx} 
\usepackage{natbib}  
\usepackage{caption} 
\usepackage{algorithm}
\usepackage{algorithmic}

\usepackage{booktabs} 
\usepackage{amsmath}
\usepackage{newfloat}
\usepackage{listings}
\DeclareCaptionStyle{ruled}{labelfont=normalfont,labelsep=colon,strut=off} 
\floatstyle{ruled}
\newfloat{listing}{tb}{lst}{}
\floatname{listing}{Listing}
\nocopyright

\title{Personalized Knowledge Tracing through Student Representation Reconstruction and Class Imbalance Mitigation}
\author {
    Zhiyu Chen\textsuperscript{\rm 1}, 
    Wei Ji\textsuperscript{\rm 1}, 
    Jing Xiao\textsuperscript{\rm 1}\thanks{Corresponding author.},
    Zitao Liu\textsuperscript{\rm 2}
}
\affiliations {
    \textsuperscript{\rm 1}School of Computer Science, South China Normal University\\Guangzhou, China, 510631\\
   
    \textsuperscript{\rm 2}Guangdong Institute of Smart Education, Jinan University\\
   Guangzhou, China, 510610\\
}

\usepackage{bibentry}

\begin{document}

\maketitle

\begin{abstract}
Knowledge tracing is a technique that predicts students’ future performance by analyzing their learning process through historical interactions with intelligent educational platforms, enabling a precise evaluation of their knowledge mastery. Recent studies have achieved significant progress by leveraging powerful deep neural networks. These models construct complex input representations using questions, skills, and other auxiliary information but overlook individual student characteristics, which limits the capability for personalized assessment. Additionally, the available datasets in the field exhibit class imbalance issues. The models that simply predict all responses as correct without substantial effort can yield impressive accuracy. In this paper, we propose PKT, a novel approach for personalized knowledge tracing. PKT reconstructs representations from sequences of interactions with a tutoring platform to capture latent information about the students. Moreover, PKT incorporates focal loss to improve prioritize minority classes, thereby achieving more balanced predictions. Extensive experimental results on four publicly available educational datasets demonstrate the advanced predictive performance of PKT in comparison with 16 state-of-the-art models. To ensure the reproducibility of our research, the code is publicly available at https://anonymous.4open.science/r/PKT.
\end{abstract}

\section{Introduction}\label{sec1}
\begin{figure}[t]
\centering
\includegraphics[width=0.9\columnwidth]{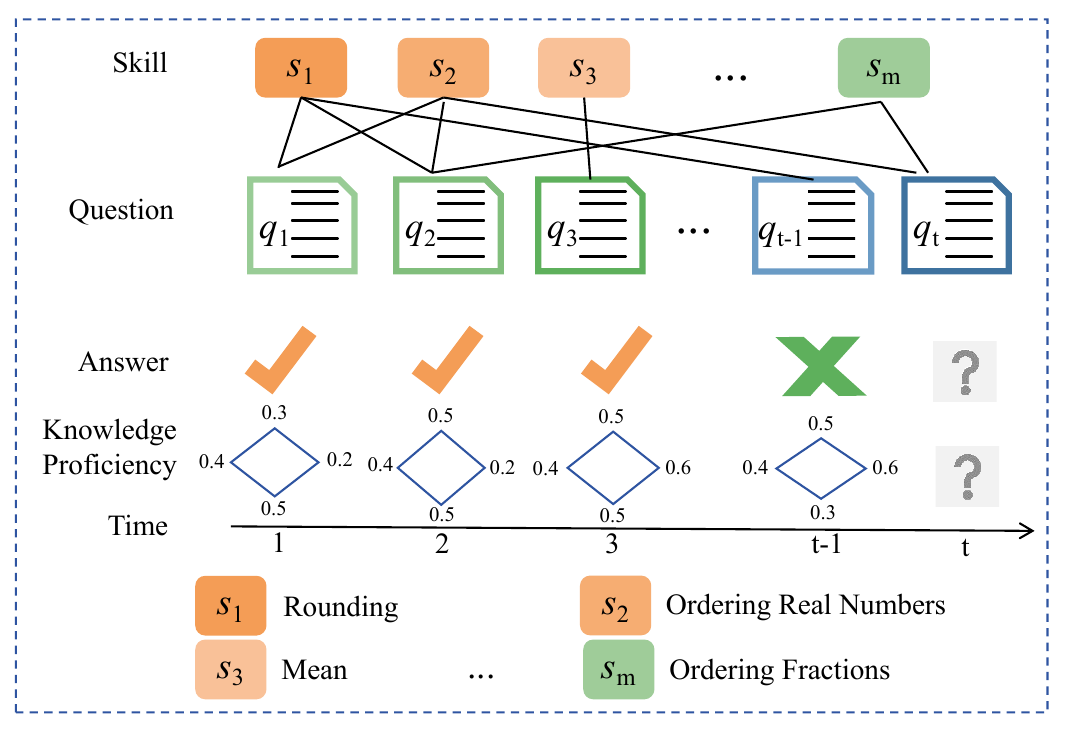} 
\caption{Illustration of learning scenarios within knowledge tracing tasks.}
\label{fig1}
\end{figure}

Knowledge tracing (KT) is a continuous predictive task aimed at simulating students' performance on questions by establishing models to forecast their level of knowledge mastery during interactions with learning platforms. This process leverages students' historical learning interaction data to construct models for estimating their proficiency levels and utilizes these models to predict their performance over a future period of time. Figure ~\ref{fig1} shows an illustrative example of the KT task. KT holds significant potential to support educators in identifying students who require additional attention, recommending tailored learning materials, and delivering valuable feedback to enhance student learning outcomes. Additionally, KT can be utilized to customize learning plans, provide early warnings, and offer targeted guidance in instructional practices, thereby enhancing learning outcomes and student performance.

\begin{figure}[t]
\centering
\includegraphics[width=1\columnwidth]{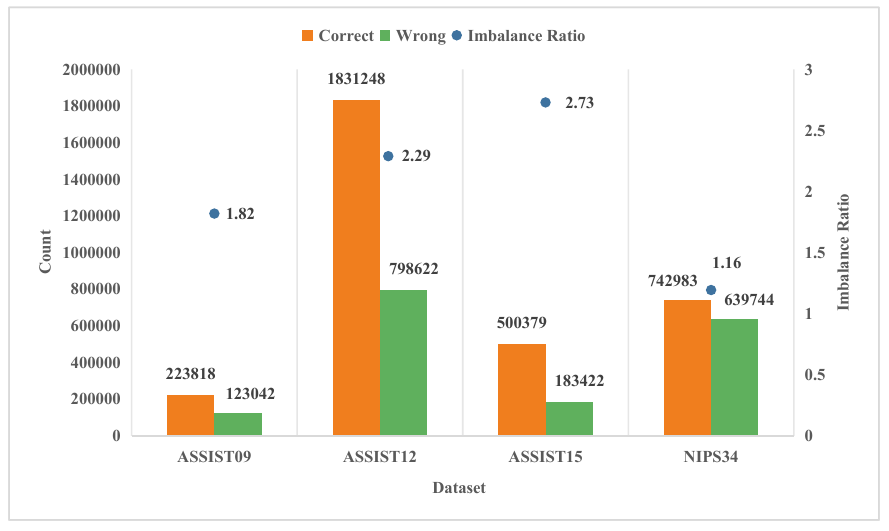} 
\caption{Class imbalance ratios in publicly available datasets within knowledge tracing.}
\label{fig2}
\end{figure}

KT has been a prominent research field since the 1990s. Its inception traces back to the pioneering work of Corbett and Anderson, who are the first to attempt estimating students' current knowledge regarding individual knowledge components (KCs) \cite{at:94}. With the advancement of deep learning techniques, deep learning-based knowledge tracing (DLKT) models have emerged, encompassing various types such as autoregressive \cite{pc:18,xp:21,kk:19,ck:18}, memory-augmented \cite{ga:19,sh:21,jz:17}, and attention-based \cite{ag:20,sl:20,sp:20,mz:21}. As DLKT models evolve, researchers continuously explore novel approaches to enhance predictive capabilities, including integration of learning-related information such as question texts, question similarities, and question difficulties. This diversification in model design renders KT applications in education more adaptable and effective.

Despite the previous success of DLKT methods in accuracy performance and the exploration of interpretability, they still exhibit limitations. On one hand, the student records provided by real educational environments mainly involve information such as students, questions, skills, correctness, and some ancillary data. However, most models primarily focus on exploring question or skill levels and their interrelations. They often neglect to investigate the underlying information about the students themselves \cite{hs:20}. Consequently, personalized KT predictions cannot be achieved. On the other hand, publicly available data collected by educational platforms suffer from class imbalance issues. Specifically, student responses are often categorized as either correct or wrong, commonly denoted by 0 or 1. Through the analysis of four commonly used public benchmark datasets within the domains, as illustrated in Figure ~\ref{fig2}, we observe imbalance ratios of 1.82, 2.29, 2.73, and 1.16, respectively. This implies that models do not necessarily require significant advantages; simply predicting all student responses as correct could yield impressive accuracy.

\begin{figure*}[t]
\centering
\includegraphics[width=1.8\columnwidth]{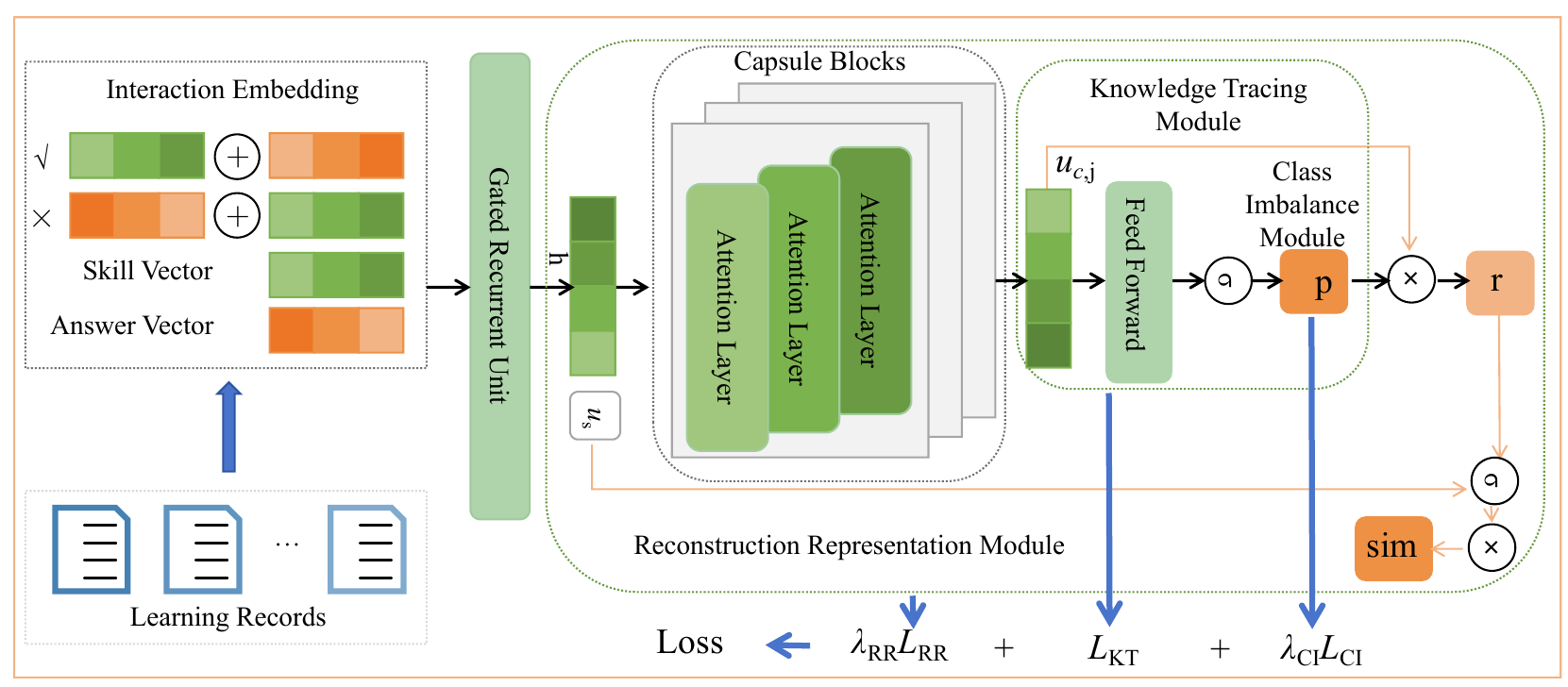} 
\caption{An overview of the proposed PKT model. $h$, $u_s$, and $u_{c,j}$ represent the hidden state, student representation, and capsule representation, respectively. $p$, $r$, and $sim$ correspond to performance prediction, reconstructed student representation, and similarity. $\sigma$, $+$, and $\times$ denote the sigmoid, concatenation and multiplication functions, respectively.}
\label{fig3}
\end{figure*}

In this paper, in order to cope with the KT problem mentioned above, we propose an innovative model, called PKT. This model leverages the integration of reconstruction and balancing modules with traditional KT, thereby enhancing personalized assessment and effectively addressing class imbalance issues. Specifically, in the task of KT, data records generated from interactions between students and educational platforms can effectively represent students themselves. In other words, our focus lies in the task of KT, which involves assessing students' proficiency based on their historical practice records and interaction information to predict their ability to correctly answer future questions. Therefore, attributes derived from student-platform interactions in the context of KT tasks can be reconstructed as representations of students. Additionally, the balancing module primarily addresses class imbalance issues regarding the correctness attribute in the dataset. Inspired by focal loss, we introduce the concept of class hardness to focus more attention on predicting minority class samples, thereby achieving a more balanced model. 

The primary innovations of our research are summarized as follows:

\begin{itemize}
\item We personalize the assessment of students' knowledge mastery by reconstructing student representations from records of interactions between students and educational scenarios, rather than constructing complex input representations using attributes provided by educational platforms.

\item We address the issue of class imbalance in the field of knowledge tracing, emphasizing predictions for minority classes to achieve a more balanced model.

\item We validate the effectiveness of the model across four public datasets by comparing it with 16 existing deep knowledge tracing models. Furthermore, we conduct comprehensive quantitative and qualitative analyses on model parameters and visualize relevant attention weights.
\end{itemize}

\section{Related Work}\label{sec2}
In this section, we first delineate relevant KT works. Then, we present a brief overview of class balance issue and focal loss.

\subsection{Knowledge Tracing}
As a result of the expansion of deep learning methods over the past decade, researchers have been trying to incorporate deep learning strategies into KT research, contributing significantly to the development of online education. 

 The pioneering work of Deep Knowledge Tracing (DKT), which applies deep learning to the field of KT, has sparked a series of methods primarily focused on sequence modeling \cite{cp:15}. DKT+ enhances the basic DKT loss function by introducing two additional regularization terms to address the limitations in reconstructing response inputs and reducing inconsistency in predicting exercises with similar concepts \cite{ck:18}. 
 
 Inspired by memory-augmented neural networks, researchers have enhanced DKT by introducing an external memory structure to better track students' learning of complex concepts with stronger representational capacity. Specifically, the knowledge state is represented using a key-value memory structure, where the key matrix stores concept representations and the value matrix stores the degree of mastery for each concept by the student \cite{jz:17}. SKVMN addresses irrelevant knowledge concepts in DKT and DKVMN by using an enhanced Hop-LSTM, while retaining DKVMN's key-value memory structure and loss function. \cite{ga:19}.

 In addressing the lack of interpretability in DKT, researchers have attempted to embed interpretability directly into the structure of individual models. The SAKT model, for the first time, models the interactive embedding sequences using a self-attention mechanism to learn the importance of each practice \cite{slp:19}. The SAINT model tackles the shallow attention layer issue of SAKT by enhancing performance through the addition of attention layers and increasing the number of layers in the encoder and decoder \cite{yc:20}. AKT incorporates a context-aware attention mechanism, considering students' overall interaction history, the influence of past questions and responses, and quantifying the time difference between previously answered questions \cite{ag:20}. 

\subsection{Class Balance and Focal Loss}

Deep neural networks, leveraging the powerful representational capabilities learned from high-quality data, have been successfully applied to various tasks \cite{yfz:23}. However, class imbalance often arises when sample distribution is uneven, particularly in educational contexts where varying knowledge component difficulties lead to disproportionate practice samples. This imbalance can cause models to overemphasize data-rich classes, impairing performance and generalization in KT tasks. Focal loss, which is widely used in tasks such as object detection and text classification, addresses this issue by down-weighting easily classified samples, enabling the model to focus on more challenging samples. In KT tasks, focal loss improves learning efficiency and mitigates bias towards majority classes, thereby enhancing the model's ability to handle class imbalance \cite{ty:17}.

\section{Problem Statement}

KT endeavors to monitor the evolution of students' cognitive states throughout the learning process, while leveraging their prior interactions to forecast their future performance on subsequent tasks. It can be succinctly articulated as follows: Given the historical sequence of student interactions, denoted as $X = \{x_1, x_2, ... , x_t\}$, where each interaction $ x = \left \langle q, \{s \vert s \in N_{q}\}, a, t \right \rangle $ signifies the student's response $a$ to question $q$ involving a set of skills $s$ at time $t$, where $a \in \{0,1\}$ is a binary indicator denoting correctness ($1$ for correct, $0$ for wrong), the aim is to predict the probability of the student answering the question correctly at time $t + 1$.

\section{The PKT Model}\label{sec3}

The architecture of our method, shown in Figure ~\ref{fig3}, includes five components: (1) the student representation module uses GRU to encode skill and response information from historical practice records; (2) the capsule blocks module creates capsule representations via an attention mechanism; (3) the knowledge tracing module predicts the probability $p$ of correctly answering the next question using a sigmoid function; (4) the reconstruction module calculates the reconstruction representation by multiplying $p$ with the student’s representation; (5) the class imbalance module uses focal loss to reduce the influence of simple examples, focusing on challenging cases.

\subsection{Student Representation Module}
Inspired by the simple but effective DKT, each interaction is initially characterized as $e$ through the encoding of skills and the information provided in responses:
\begin{equation}
    e_{t} = s_{t} + a_{t} \times E
\end{equation}
where $E$ is the total number of skills.

Standard RNNs struggle with long-term dependencies. The Gated Recurrent Unit (GRU) addresses this issue with its simplicity, fewer parameters, faster training, and superior performance across tasks. A GRU cell controls the flow of information through two gating mechanisms to compute the hidden state $h_t$. Given the input $e_t$ at the current time step $t$ and the hidden state $h_{t-1}$ at the previous time step $t-1$, the calculations for the update gate $z_t$, reset gate $r_t$, candidate hidden state $\tilde{h}_{t}$ and the final hidden state $h_t$ are as follows:
\begin{align}
    z_t = \sigma(W_z [h_{t-1}, e_t] + b_z),\nonumber \\
    r_t = \sigma(W_r [h_{t-1}, e_t] + b_r), \nonumber
    \\
    \tilde{h}_{t} = tanh(W_h [r_t \odot h_{t-1},e_t] + b_h), \nonumber
    \\
    h_t = (1-z_t) \odot h_{t-1} + z_t \odot \tilde{h}_{t}
\end{align}
where $W_z$, $W_r$, and $W_h$ are weight matrices, $b_z$, $b_r$, and $b_h$ are bias vectors, $\sigma$ is a Sigmoid function. $\odot$ denotes element-by-element multiplication, and $[h_{t-1}, x_t ]$  indicates stitching the hidden state $h_{t-1}$ of the previous time step with the input $x_t$ of the current time step. 

Formally, the student representation $u_s$ is the average of the hidden vectors obtained from GRU:
\begin{equation}
    u_s = \frac{1}{N_s} \sum_{j=1}^{N_s} h_i
\end{equation}
where $N_s$ denotes the actual number of practice questions in the student's sequence, and each item is represented by a dense vector.

\begin{table}[ht]
    \centering
	\begin{tabular}{@{}lllll@{}}
		\toprule    
		Items&ASSIST09&ASSIST12&ASSIST15&NIPS34\\
		\midrule
		\# user&4,661&33,568&19,292&9,401\\
		\# question&17,737&53,070&-&948\\
		\# skill&123&265&100&57\\
		\# record&337,415&2,709,568&682,789& 1,399,470\\
		\# maxlen&88&99&36&285\\
            \# s per q&1.197&1&-&1\\
            \# max skill&4&1&1&2\\
            \# ratio&1.82&2.29&2.73&1.16\\
		\bottomrule
	\end{tabular}
    \caption{Data statistics from four widely available knowledge tracing datasets.}\label{tab1}%
\end{table}

\begin{table*}[ht]
       \begin{tabular*}{\textwidth}{@{\extracolsep{\fill}}lcccccccc@{\extracolsep{\fill}}}
		\toprule%
		& \multicolumn{2}{@{}c@{}}{ASSIST09} & \multicolumn{2}{@{}c@{}}{ASSIST12} & \multicolumn{2}{@{}c@{}}{ASSIST15} & \multicolumn{2}{@{}c@{}}{NIPS34} \\\cmidrule{2-3}\cmidrule{4-5}\cmidrule{6-7}\cmidrule{8-9}%
		Methods & AUC &ACC &AUC &ACC &AUC &ACC &AUC &ACC\\
		\midrule
DKT&0.74970971&0.72051900&0.73064160&0.7339984&0.72490002&0.75053495&0.76655805&0.70169442\\
DKT+&0.75119459&0.72245187&0.73122078&0.73372978&0.72740154&0.75115162&0.76612178&0.70131427\\
DKT-F&-&-&0.73204137&0.73512508&-&-&0.76506680&0.7007828\\
KQN&0.69934021&0.65786351&0.72794852&0.73303400&0.72456062&0.75008172&0.76321956&0.69847608\\
DKVMN&0.74236621&0.71772495&0.72360976&0.73149320&0.72124936&0.75005200&0.75446065&0.69062256\\
Deep-IRT&0.74282364&0.71818425&0.72649645&0.73267958&0.72112423&0.72112423&0.76267809&0.69728027\\
SKVMN&0.72049684&0.71116086&0.69071281&0.71881429&0.70325808&0.74433101&0.72472423&0.66620409\\
SAKT&0.72217767&0.70329544&0.70606223&0.72329677&0.69745003&0.74504428&0.74660217&0.68357070\\
ATKT&0.73709918&0.71496918&0.72536719&0.73110334&0.72001504&0.74670114&0.75587901&0.69168509\\
SAINT&0.69329369&0.69365024&0.66977556&0.71347747&0.65200983&0.73427098&0.78021620&0.71241239\\
AKT&0.78406651&0.74141692&\textbf{0.78018800}&0.75823513&0.72609858&0.75167171&0.79947800&0.72862958\\
HawkesKT&0.73002914&0.70660619&0.74825609&0.74029588&-&-&0.60028020&0.61411299\\
DIMKT&0.76334256&0.72920733&0.76978531&0.75115606&-&-&0.80016017&0.72945262\\
AT-DKT&0.75012817&0.72094002&0.73510385&0.73626482&-&-&0.77698865&0.71145648\\
simpleKT&0.77364423&0.72702568&0.77343904&0.75195630&0.72335022&0.75042350&0.79927761&0.72924963\\
SparseKT&0.76883591&0.72629846&0.77119637&0.75450999&0.72239032&0.75052752&0.79654053&0.72599069\\
		\midrule	
PKT&\textbf{0.80328128}&0.75518456&0.77765480&0.75397372&\textbf{0.76066066}&0.74736508&\textbf{0.80199695}&0.72602759\\
		\bottomrule
	\end{tabular*}
    \caption{Performance comparison on four benchmark datasets.}\label{tab2}%
\end{table*}

\subsection{Capsule Blocks Module}
Capsule blocks are used to construct the capsule structure, one for the `correct' activation state and the others for the `wrong' activation states. Given the output vectors encoded by GRU (i.e., $h_t$), the attention mechanism is used to construct capsule representations:
\begin{align}
    a_{t,j} = h_t W_{a,j},\nonumber \\
    \alpha_{t,j} = \frac{exp(a_{t,j})}{\sum_{i=1}^{N_s}  exp(a_{i, j})},\nonumber \\
    u_{c, j} = \sum_{i=1}^{N_s} \alpha_{t,j} h_t,\nonumber \\
\end{align}

where $W_{a,j}$ represents the parameter of capsule block $j$ of the attention layer. The attention score for each position, $a_{t,j}$, is derived by multiplying the representation $h_t$ by the weight matrix $W_{a,j}$ and subsequently normalizing it to form a probability distribution across the items $\alpha_{t,j} =[\alpha_{1,j}, \alpha_{2,j}, ..., \alpha_{N_s,j}]$. Finally, the capsule representation vector, $u_{c,j}$, is obtained as a weighted summation of all positions using the attention scores as weights.

\subsection{Knowledge Tracing Module}
We use the capsule representation vector obtained from the capsule blocks module to calculate the probability $p$ that a student can correctly answer the next question:
\begin{align}
    p_j = \sigma(W_{p,j} u_{c,j} + b_{p,j}),\nonumber \\
    p = mean(stack(\sum_{j=1}^{N_c} p_j))
\end{align}
    
where $W_{p,j}$ and $b_{p,j}$ are the probability parameters for the current capsule block j, $N_c$ represents the number of capsule blocks.

\subsection{Reconstruction Representation Module}
Note that the capsule representation vector $u_{c, j}$ acquired through the attention mechanism constitutes a sophisticated encoding of the complete input student sequence information. This vector will be employed for the student representation reconstruction. Such reconstruction involves multiplying $u_{c, j}$ by the probability $p_j$:
\begin{align}
    r_{s,j} = p_j u_{c,j},\nonumber \\
    r = max(stack(\sum_{j=1}^{N_c} r_{s,j}))
\end{align}

To assess the effectiveness of the reconstructed representation, we evaluate the similarity between the reconstructed representation and the original student representation, primarily using the inner product between the reconstructed representation and the original student representation:
\begin{equation}
    sim = \sigma (u_s r)
    \label{eq7}
\end{equation}

\subsection{Class Imbalance Module}
The cross-entropy loss function is commonly utilized in the training of DLKT models. However, the phenomenon of class imbalance often results in easily classifiable instances dominating the loss and gradient computations. Drawing inspiration from the superior performance of focal loss in addressing long-tail issues in the visual domain, we apply focal loss to KT tasks. Focal loss not only balances the importance of positive and negative examples but also distinguishes between easy and hard examples. Specifically, it reduces the weight of easily classifiable examples, thereby emphasizing the training on hard-to-classify negative instances:
\begin{equation}
    L_{CI} = - \alpha_{CI} (1 - p)^\gamma \log(p)
\end{equation}

where $\alpha_{CI}$ is the weighting factor, empirically set to the class imbalance ratio, and $\gamma$ is the tunable focusing parameter, $\gamma \ge 0 $.

\subsection{Training Objective}
The training goal of the PKT model involves minimizing performance prediction loss, reconstruction loss of student representations, and addressing class imbalance. To achieve these goals, the model incorporates two additional losses: reconstruction loss and class balance loss, which complement the primary KT task.

The basic KT task still employs gradient descent to update the parameters of the model, aiming to minimize the cross-entropy loss between the model's final predictions and the ground truth labels:
\begin{equation}
    L_{KT} = \sum_{t=1}^{T} (\alpha_{KT} \log(p) + (1 - \alpha_{KT}) \log(1 - p))
\end{equation}
where T is the maximum length of the student sequence.

To ensure the similarity between the reconstructed representation obtained through the capsule blocks module and the original student representation, similarly, we utilize the previously obtained similarity $sim$ with the ground truth labels for cross-entropy loss computation:
\begin{equation}
    L_{RR} = \sum_{t=1}^{T} (\alpha_{RR} \log(sim) + (1 - \alpha_{RR}) \log(1 - sim))
\end{equation}

Finally, the overall training objective will be achieved through the following formula:
\begin{equation}
    L = L_{KT} + \lambda_{RR} L_{RR} + \lambda_{CI} L_{CI}
\end{equation}
where $\lambda_{RR}$ and $\lambda_{CI}$ are hyperparameters.

\section{EXPERIMENT} \label{sec4}
In this section, we first experiment with four publicly educational datasets to comprehensively evaluate the effectiveness and advantage of the PKT. 

\subsection{Datasets}
There are four publicly available datasets \cite{f:09,s:20} that support research on KT tasks. Table ~\ref{tab1} lists their statistical information.

Following the preprocessing method from PYKT \cite{zt:22}, we perform these steps: (1) remove attributes with null values; (2) exclude students with fewer than three records; (3) split questions involving multiple skills into separate interactions while retaining the `user\_id'; (4) adjust sequence lengths to the average student sequence length. Sequences longer than this average are truncated, those between three and the average are padded with -1, and sequences with fewer than three records are discarded.

\begin{table*}[ht]
	\begin{tabular*}{\textwidth}{@{\extracolsep{\fill}}lccccccccc@{\extracolsep{\fill}}}
		\toprule%
		& \multicolumn{2}{@{}c@{}}{ASSIST09} & \multicolumn{2}{@{}c@{}}{ASSIST12} & \multicolumn{2}{@{}c@{}}{ASSIST15} & \multicolumn{2}{@{}c@{}}{NIPS34} \\\cmidrule{2-3}\cmidrule{4-5}\cmidrule{6-7}\cmidrule{8-9}%
		Methods & AUC &ACC &AUC &ACC &AUC &ACC &AUC &ACC\\
		\midrule
		PKT-RR& 0.77398075&0.73589047&0.73390315&0.73236272&0.72335068
 & 0.75738344&0.78592351&0.71777506\\
		PKT-CI&0.76300987&0.72110218&0.72658700&0.72730048&\textbf{0.76074511}
 &0.74540069&0.77616061&0.71000970\\
		PKT-RR\&CI&0.72215882 &0.70540889&0.71126169&0.72535952
&0.69137712&0.75266136&0.73879955&0.68312252\\
		PKT&\textbf{0.80328128}&0.75518456&\textbf{0.77765480}&0.75397372
 &0.76066066& 0.74736508&\textbf{0.80199695}&0.72602759\\
		\bottomrule
	\end{tabular*}
    \caption{Contribution analysis of representation reconstruction and class balancing.}\label{tab3}
\end{table*}

\begin{figure*}[t]
\centering
\includegraphics[width=2\columnwidth]{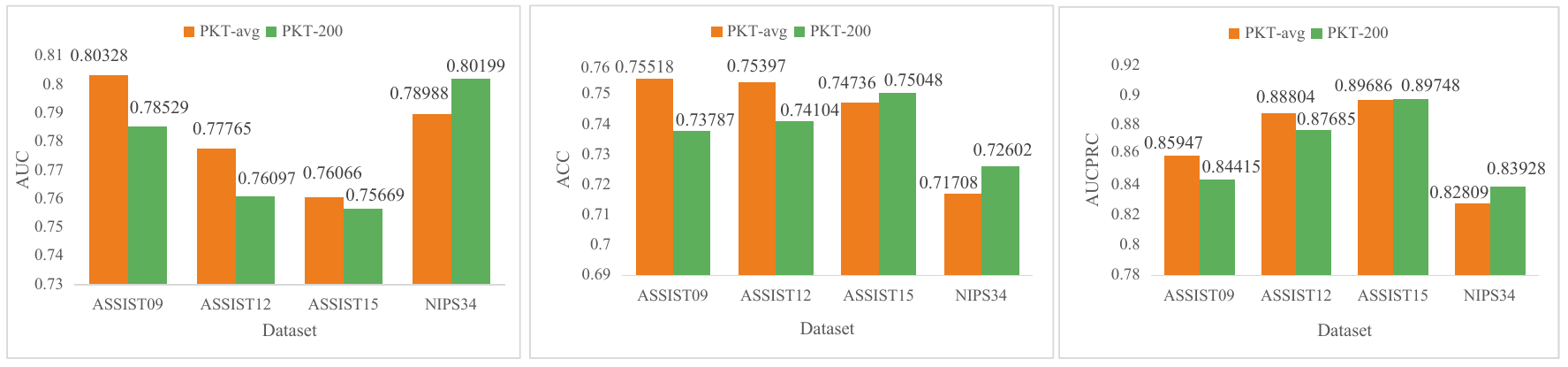} 
\caption{Performance comparison between setting maxlen to the average student sequence length and a fixed value of 200.}
\label{fig4}
\end{figure*}

\subsection{Baselines}
To validate the superiority and effectiveness of our model, we evaluate a total of 16 representative benchmark works. In addition to the aforementioned related works, we further compare the PKT model with the following state-of-the-art DLKT models including KQN \cite{jl:19}, Deep-IRT \cite{ck:19}, HawkesKT \cite{cyw:21}, DIMKT \cite{shs:22}, AT-DKT \cite{ztl:23}, SimpleKT \cite{tl:23}, and SparseKT \cite{syh:23}.

\subsection{Experimental Setup} 
Following PYKT, we use 5-fold cross-validation for all DLKT methods and datasets. Interaction sequences are split with 80\% for training and validation and 20\% for testing. The maximum sequence length is set to the average for each dataset. We train the model with the ADAM optimizer for 200 epochs, applying early stopping if AUC does not improve in 10 epochs. All experiments are run on an NVIDIA GeForce RTX 3090. AUC serves as the primary metric, while accuracy and AUCPRC (Area Under the Precision-Recall Curve) are additionally employed to evaluate performance and ensure class balance \cite{jd:06}.

\subsection{Results} 

\subsubsection{Overall Performance}
Table ~\ref{tab2} summarizes the predictive performance of PKT and all baseline works on four publicly available datasets in terms of both AUC and ACC metrics. We can observe the following results: (1) PKT significantly outperforms the 16 baselines on all four datasets (except for AKT on ASSIST12 dataset which has a 0.25\% AUC loss). More importantly, as a representative of DLKT models, our proposed model improves the AUC by 1.92\%, 3.46\%, and 0.25\% on the ASSIST09, ASSIST15, and NIPS34 datasets, respectively, compared to AKT. This overall validates the effectiveness and superiority of our approach. (2) DKT is still a very reliable baseline across all datasets compared to its variants such as KQN, DKVMN, DeepIRT, and even some attention-based models, for instance, SKVMN, SAKT, and ATKT. DKT has better AUC performance compared to KQN by 0.69\%, 0.27\%, 0.03\%, and 3.32\% on ASSIST09, ASSIST12, ASSIST15, and NIPS34 datasets, respectively. (3) SAINT performs poorly on all the ASSIST datasets we used, potentially due to the sparser nature of these datasets compared to EDnet, which was used in the original paper. However, SAINT outperforms DKT and its variants on the NIPS34 dataset. We hypothesize that this is because NIPS34 has a longer average sequence length than the ASSIST datasets, allowing SAINT to more effectively capture historical interaction information through the attention mechanism. (4) Despite using only skill-related information, PKT performs better than models that simultaneously use both question and skill information, such as SimpleKT, HawkesKT, DIMKT, AT-DKT, and SparseKT.

\begin{figure}[t]
\centering
\includegraphics[width=0.9\columnwidth]{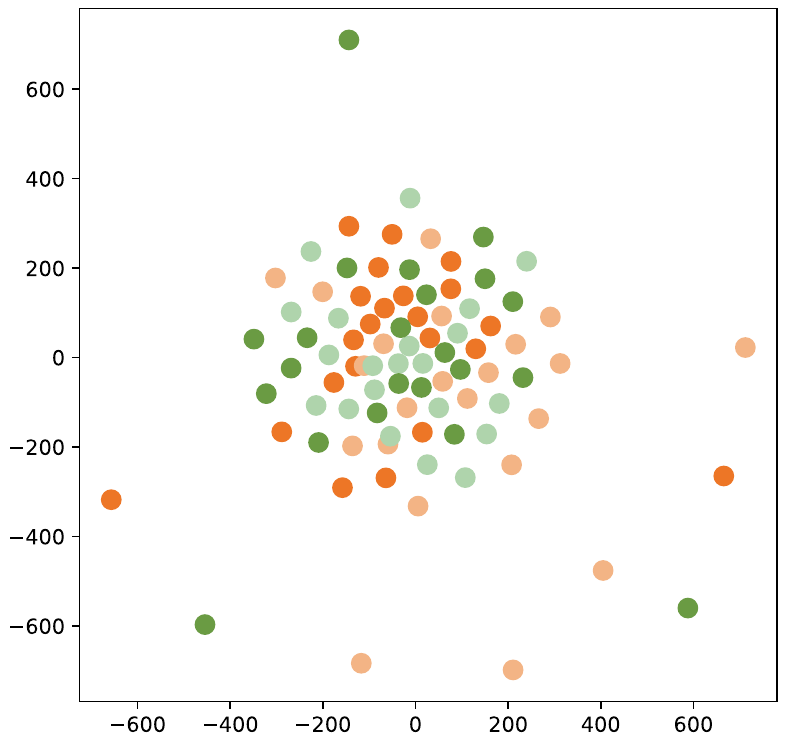} 
\caption{Visualization of the reconstructed and original student representations using t-SNE, showing clustering of similar representations and dispersion of outliers.}
\label{fig5}
\end{figure}

\begin{figure*}[ht]%
	\centering
	\includegraphics[width=1\textwidth]{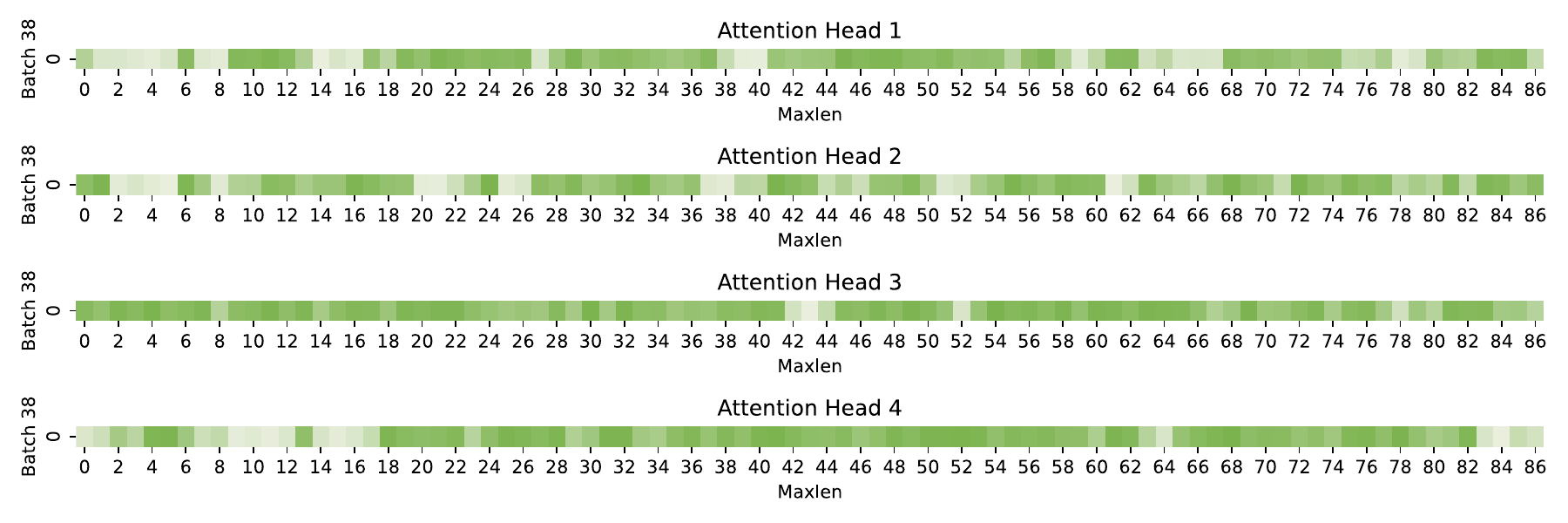}
	\caption{Visualization of PKT’s prediction results with attention weights from four heads, representing different spatial information of historical interactions.}\label{fig6}
\end{figure*}

\begin{table}[ht]
    \centering
	\begin{tabular}{@{}lllll@{}}
		\toprule    
		Method&ASSIST09&ASSIST12&ASSIST15&NIPS34\\
		\midrule
		PKT-RR &0.83531 &0.85999 &0.87512 &0.82619 \\
            PKT-CI &0.83531 &0.85581 &0.89430 &0.81673 \\
            PKT-RR\&CI &0.79958 &0.84267 &0.85443 &0.77661\\
		\midrule	
		PKT  &\textbf{0.85947} &\textbf{0.88804} &\textbf{0.89686} &\textbf{0.83928} \\
		\bottomrule
	\end{tabular}
    \caption{Comparative analysis of ablation experiments on AUCPRC}\label{tab4}%
\end{table}

\subsubsection{Ablation Study}
To validate the effectiveness of the key components in the PKT model, we conduct three sets of ablation experiments to compare different variants of PKT. Specifically, PKT-RR denotes the PKT model excluding student reconstruction information; PKT-CI represents the PKT model under class-imbalance conditions; and PKT-RR\&CI refers to the absence of both reconstruction information and class imbalance, as depicted in Table ~\ref{tab3}. From Table ~\ref{tab3}, it is evident that: (1) PKT outperforms all three model variants and is comparable to PKT-CI on the ASSIST 2015 dataset. This empirically confirms the importance of reconstructing student representations and maintaining model balance across classes in predicting student performance; (2) When comparing PKT-RR\&CI with PKT-RR and PKT-CI, we find that the enhancements from reconstructing student information and balancing classes are complementary.

The Area Under the Precision-Recall Curve (AUCPRC) assesses a model's ability to recognize positive samples by evaluating the relationship between precision and recall at various threshold levels. Compared to AUC, AUCPRC more effectively reflects a model's performance in detecting rare events or minority classes, making it widely applicable in fields such as medical diagnostics and anomaly detection. A high AUCPRC value indicates that a model can maintain a high recall rate while achieving a high precision rate. Inspired by this, we apply AUCPRC to the domain of KT, reassessing PKT on imbalanced public datasets, as demonstrated in Table ~\ref{tab4}. Notably, on the ASSIST15 dataset, PKT achieves a higher AUCPRC than all its variants, validating the effectiveness of our introduction of focal loss in addressing class imbalance in publicly available datasets within KT domain.

\subsubsection{ Analysis of sequence length}
 The key distinction from the preprocessed data in PYKT is that, for each dataset, the sequence length is set to the average sequence length of the student rather than a fixed length of 200. As shown in Figure ~\ref{fig4}, setting the maximum sequence length to the student's average length improves PKT's performance by 1.80\%, 1.67\%, and 0.40\% on the ASSIST09, ASSIST12, and ASSIST15 datasets, respectively. However, it results in a 1.21\% performance decline on the NIPS34 dataset. This decline on the ASSIST datasets occurs because the average sequence length is less than 200, which means that using a fixed length introduces many -1 values as padding, adding invalid information. Overall, this validates the effectiveness of setting PKT's input sequence length to the average student sequence length for each dataset.

\subsubsection{Visualization of Similarity}

  As shown in Figure~\ref{fig5}, we randomly select a sequence record of a student's practice skill in the ASSIST09 dataset and calculate the similarity between the reconstructed representation and the original student representation using the inner product, denoted as $sim$ in Eq.~\ref{eq7}. We then employ the t-SNE algorithm to project the representations into a two-dimensional space, with the x-axis and y-axis representing two new feature dimensions. These dimensions do not directly correspond to the original data's features but aim to preserve the similarity between data points. The use of t-SNE for dimensionality reduction enables us to visualize how the representations cluster and interact in a two-dimensional plane. Similar representations tend to cluster together, while dissimilar ones remain farther apart. In Figure~\ref{fig5}, most data points cluster together, except for a few outliers, which confirms that the reconstructed representations are similar to the original student representations.

  \subsubsection{Visualization of Attention}

 In this section, we qualitatively present the visualization of the prediction results generated by PKT, as shown in Figure ~\ref{fig6}. To better understand the predictive behavior of the model, we randomly select the weights of the associations between the predicted future performance of the user and their historical interactions recorded in the ASSIST2009 dataset. The model utilizes four attention heads, each focusing on different aspects of the historical interaction sequence, thereby enhancing the overall prediction accuracy of the model.


\section{Conclusion} \label{sec5}
In this paper, we present a personalized knowledge tracing (PKT) approach. In contrast to recent DLKT models, PKT reconstructs student representations from historical interaction sequences rather than constructing complex inputs from questions, skills, and auxiliary information. In addition, PKT employs focal loss to direct student attention to minorities, allowing the model to assess students' abilities in more balanced scenarios. Through qualitative and quantitative analyses, PKT outperforms 16 state-of-the-art DLKT methods across four publicly available datasets, achieving superior AUC and AUCPRC, with AUCPRC being a common metric for class imbalance issues.

\bibliography{aaai25}
\bibliographystyle{aaai25}

\section*{Reproducibility Checklist}

This paper:
\begin{itemize}
\item  Includes a conceptual outline and/or pseudocode description of AI methods introduced: [yes]
\item  Clearly delineates statements that are opinions, hypothesis, and speculation from objective facts and results: [yes]
\item  Provides well marked pedagogical references for less-familiare readers to gain background necessary to replicate the paper: [yes]
\end{itemize}
Does this paper make theoretical contributions? [yes]

If yes, please complete the list below.
\begin{itemize}
\item  All assumptions and restrictions are stated clearly and formally: [yes]
\item  All novel claims are stated formally (e.g., in theorem statements): [yes]
\item  Proofs of all novel claims are included: [yes]
\item  Proof sketches or intuitions are given for complex and/or novel results: [yes]
\item  Appropriate citations to theoretical tools used are given: [yes]
\item  All theoretical claims are demonstrated empirically to hold: [yes]
\item  All experimental code used to eliminate or disprove claims is included: [yes]
\end{itemize}

Does this paper rely on one or more datasets? [yes]

If yes, please complete the list below.
\begin{itemize}
\item  A motivation is given for why the experiments are conducted on the selected datasets:  [yes]
\item  All novel datasets introduced in this paper are included in a data appendix:  [NA]
\item  All novel datasets introduced in this paper will be made publicly available upon publication of the paper with a license that allows free usage for research purposes: [yes]
\item  All datasets drawn from the existing literature (potentially including authors’ own previously published work) are accompanied by appropriate citations: [yes]
\item  All datasets drawn from the existing literature (potentially including authors’ own previously published work) are publicly available: [yes]
\item  All datasets that are not publicly available are described in detail, with explanation why publicly available alternatives are not scientifically satisficing: [NA]
\end{itemize}

Does this paper include computational experiments? [yes]

If yes, please complete the list below.

\begin{itemize}
\item  Any code required for pre-processing data is included in the appendix: [yes] All codes can be found at https://anonymous.4open.science/r/PKT, as mentioned in the abstract.
\item  All source code required for conducting and analyzing the experiments is included in a code appendix: [yes] All codes can be found at https://anonymous.4open.science/r/PKT. as mentioned in the abstract.
\item  All source code required for conducting and analyzing the experiments will be made publicly available upon publication of the paper with a license that allows free usage for research purposes: [yes] 
\item  All source code implementing new methods have comments detailing the implementation, with references to the paper where each step comes from: [yes] 
\item  If an algorithm depends on randomness, then the method used for setting seeds is described in a way sufficient to allow replication of results: [yes] 
\item  This paper specifies the computing infrastructure used for running experiments (hardware and software), including GPU/CPU models; amount of memory; operating system; names and versions of relevant software libraries and frameworks: [yes] 
\item  This paper formally describes evaluation metrics used and explains the motivation for choosing these metrics: [yes] 
\item  This paper states the number of algorithm runs used to compute each reported result: [yes] 
\item  Analysis of experiments goes beyond single-dimensional summaries of performance (e.g., average; median) to include measures of variation, confidence, or other distributional information: [yes] 
\item  The significance of any improvement or decrease in performance is judged using appropriate statistical tests (e.g., Wilcoxon signed-rank): [NA]
\item  This paper lists all final (hyper-)parameters used for each model/algorithm in the paper’s experiments: [NA] The hyper-parameter settings used in the code are provided at https://anonymous.4open.science/r/PKT.
\item  This paper states the number and range of values tried per (hyper-) parameter during development of the paper, along with the criterion used for selecting the final parameter setting: [NA] All parameter settings used in the code are available at https://anonymous.4open.science/r/PKT.
\end{itemize}

\end{document}